# Structured SUMCOR Multiview Canonical Correlation Analysis for Large-Scale Data

Charilaos I. Kanatsoulis, Xiao Fu, Nicholas D. Sidiropoulos, and Mingyi Hong

*Abstract*— The sum-of-correlations (SUMCOR) formulation of generalized canonical correlation analysis (GCCA) seeks highly correlated low-dimensional representations of different views via maximizing pairwise latent similarity of the views. SUMCOR is considered arguably the most natural extension of classical two-view CCA to the multiview case, and thus has numerous applications in signal processing and data analytics. Recent work has proposed effective algorithms for handling the SUMCOR problem at very large scale. However, the existing scalable algorithms cannot incorporate structural regularization and prior information – which are critical for good performance in real-world applications. In this work, we propose a new computational framework for large-scale SUMCOR GCCA that can easily incorporate a suite of structural regularizers which are frequently used in data analytics. The updates of the proposed algorithm are lightweight and the memory complexity is also low. In addition, the proposed algorithm can be readily implemented in a parallel fashion. We show that the proposed algorithm converges to a Karush-Kuhn-Tucker (KKT) point of the regularized SUMCOR problem. Judiciously designed simulations and real-data experiments are employed to demonstrate the effectiveness of the proposed algorithm.

*Index Terms*— Canonical correlation analysis (CCA), multiview CCA, SUMCOR, scalability, feature selection, cross-modality retrival

## I. INTRODUCTION

Canonical correlation analysis (CCA) is an important tool in statistics [1], [2], [3]. Classical CCA aims at extracting common low-dimensional structure of the same set of entities measured in two different high-dimensional feature spaces, e.g., image and text representations of a person. The high-dimensional representations in different feature spaces are often referred to as 'views' or 'modalities' of the entities. CCA can be considered as a multiview generalization of principal component analysis (PCA) that handles a single view. Unlike PCA of the concatenated views, CCA intrinsically ignores strong components in the views which are irrelevant to the common latent structure – this also means that CCA is naturally robust to gross outliers in any single view.

CCA/GCCA has an impressive array of applications in data mining and machine learning, including clustering [4], regression [5], outlier detection [6], natural language processing and word embedding [7], [8], [9], speech processing [10], heath care data analytics [11], genetics [12], [13], [14] and many more. Despite its long history and pervasive applications, CCA still poses computational challenges that are intriguing from an optimization point of view – especially when the dimensions of the feature spaces and the amount of data get large. The classical CCA which considers the two-view case is optimally solvable via generalized eigen-decomposition [15], [16], [17], [18]. However, when the number of views is larger than two, most formulations of multimodal CCA [also known as generalized CCA (GCCA)] are not polynomial time solvable, and thus require effective approximation algorithms [19], [8]. Even worse, both CCA and GCCA have serious scalability problems. In the eigen-decompostion based solution of CCA, a whitening step is often needed for stably computing the generalized eigen-decomposition, which takes the square root decomposition of the auto-correlation matrices of the views and inverts them. This whitening step costs a large amount of memory and flops (e.g., for a case with 100,000 features and 100,000 samples, the whitening step consumes 75 GB memory and $10^{15}$ flops), which makes it very hard to push these methods forward to handle big data. In recent years, the machine learning and signal processing communities have both noticed the scalability issues of CCA. Several publications address these issues, but mostly for the two-view case [15], [16], [17], [18], [20], [21]. Essentially, these methods all follow the seminal work of Golub et al. in the 1990s [22] to tackle the large-scale CCA problem using alternating least squares. By interpreting the process as an inexact orthogonal iteration algorithm (which is a generalization of the power iterations), one can show these algorithm converge to the desired solution.

The algorithms and analyses in [15], [16], [17], [18], [20] for the two-view case are nice and insightful, but handling the multiview case is much more challenging, since the optimization problems are very different – e.g., the two-view case is solvable but many of its multiview extensions like sum-of-correlations (SUMCOR) GCCA are NP-hard. Very recently, Fu et al. [23] proposed an efficient way to handle the SUMCOR problem, which is the first large-scale SUMCOR algorithm that scales up to truly large views. Fu et al. [7] have also considered another popular formulation of GCCA, namely, the MAX-VAR GCCA [24], [10], [11] and proposed highly scalable algorithms for it in [7].

The prior work in [23] opens a door for handling big multiview data using SUMCOR GCCA – which is arguably the most natural extension of classical two-view CCA. However, there are still many challenges remaining. The most important open problem after [23] is how to incorporate structure promoting regularization on the canonical correlation com-

C.I. Kanatsoulis and M. Hong are with the Department of ECE, University of Minnesota, Minneapolis, MN 55455, USA email: (kanat003,mhong)@umn.edu. X. Fu is with the School of EECS, Oregon State University, Corvallis, OR 97331, USA email: xiao.fu@oregonstate.edu. N. D. Sidiropoulos is with the Department of ECE, University of Virginia, Charlottesville, VA 22904, USA email: nikos@virginia.edu.

ponents in big multiview analysis. Note that taking structure promoting terms into consideration is of great interest to data analytics since it is one of the most straightforward ways of incorporating prior information. For example, in genetics [12], [13], sparse GCCA has been used extensively. Sparse GCCA imposes sparsity on the canonical correlation components (the dimensionality-reducing matrices for the different views), so that many irrelevant features of the views can be ignored when learning the correlated representations. For perfect modeling and noiseless data, this would not be necessary – but in practice sparsity helps substantially since real world modeling is never perfect and the data is usually quite noisy. Sparsity might be the most popular structure that is considered in the literature [12]; other types of structure such as nonnegativity and smoothness have also been used in different applications to produce desired latent representations. Notice that in terms of computations, the extension from 'plain GCCA' to structured GCCA is usually non-trivial, since the latter usually involves extra and nonsmooth terms (e.g., the sparsity-promoting terms) in its cost function. In fact, the work in [23] cannot be modified to cover the structured case, and thus the design of the algorithm needs to be completely reconsidered. We also note that there have been many works considering structured CCA/GCCA [25], [26], [13], [27], [28], but these algorithms were not designed for big data, and thus are not scalable.

**Contributions** In this work, we consider the structured SUMCOR GCCA problem in a big data setting. We propose an algorithmic framework that can easily handle a variety of structure promoting terms while computing canonical components of multiple big views. Our detailed contributions are as follows:

• **A Penalty-Dual Decomposition Based Algorithmic Framework for Structured SUMCOR.** We propose to employ a penalty-dual decomposition (PDD) technique [29] to 'split' the effort of tackling the already very hard manifold constraints of SUMCOR and the newly added regularizers. Our approach has an array of desired features. First, the algorithm admits lightweight updates leveraging data sparsity. Second, a variety of regularizers that are frequently used in data analytics, such as sparsity, group sparsity, elastic net, smoothness, and nonnegativity can be easily handled by our new framework. Third, the updates can be naturally distributed to different computational agents with limited communication overhead, which can further reduce the overall runtime.

• **Rigorous Convergence Analysis.** Since the structured SUMCOR formulation can involve a nonsmooth objective, and the manifold constraints are nonconvex, convergence of primal-dual algorithms applied to this problem is unclear. In this work, we apply the analytical tools of PDD to provide convergence guarantees. Specifically, we first show that a regularity condition (namely, *Robinson's condition*) is satisfied by the Karush-Kuhn-Tucker (KKT) points of the structured SUMCOR problem. Note that verifying Robinson's condition is in general hard, but fortunately the special problem structure of SUMCOR allows us to do this. Following this stepping stone, we further show that the PDD updates can ensure that the algorithm reaches a KKT point.

• **Extensive Simulations and Real Experiment.** We employ a set of judiciously designed simulations to showcase the effectiveness of the proposed algorithm under a large-scale scenario. In addition, we also apply the proposed algorithm to a large-scale cross-modality retrieval problem that involves more than twenty languages as the different views. Validation results show that the proposed algorithmic framework exhibits very promising performance

A preliminary version of part of this work has been submitted to ICASSP 2018. The ICASSP version [30] contains the basic algorithmic framework and part of the simulations. This journal version additionally includes the detailed convergence proof, more large-scale simulations, and the real-data experiment.

## II. BACKGROUND

### A. CCA and GCCA

Let us consider a two-view data set, where $\boldsymbol{Y}_1 \in \mathbb{R}^{L \times M_1}$ and $\boldsymbol{Y}_2 \in \mathbb{R}^{L \times M_2}$ are the two views – i.e., $\boldsymbol{Y}_1(\ell,:) \in \mathbb{R}^{1 \times M_1}$ and $\boldsymbol{Y}_2(\ell,:) \in \mathbb{R}^{1 \times M_2}$ are two high-dimensional representations of the entity $\ell$ in two different feature spaces (e.g., speech and image of a person). Let us assume that $\boldsymbol{X}_i = (1/\sqrt{L})(\boldsymbol{Y}_i - \boldsymbol{1}\boldsymbol{d}_i^T)$ is the scaled and centered version of the $i$th view, where $\boldsymbol{d}_i^T = (1/L)\sum_{\ell=1}^L \boldsymbol{Y}(\ell,:)$ is the sample mean of the $i$th view.

The classical two-view CCA can be expressed in the following optimization form [17], [22], [20], [31]:

$$\max_{\boldsymbol{Q}_1, \boldsymbol{Q}_2} \ \mathrm{Tr}\left(\boldsymbol{Q}_1^T \boldsymbol{X}_1^T \boldsymbol{X}_2 \boldsymbol{Q}_2\right) \tag{1a}$$

$$\text{s.t.} \ \boldsymbol{Q}_i^T \boldsymbol{X}_i^T \boldsymbol{X}_i \boldsymbol{Q}_i = \boldsymbol{I}_K, \ i=1,2, \tag{1b}$$

where $\boldsymbol{I}_K$ denotes a $K \times K$ identity matrix, $\boldsymbol{Q}_i \in \mathbb{R}^{M_i \times K}$ denotes a dimensionality-reducing matrix of view $i$, $K$ is the number of canonical components that we seek. Maximizing $\mathrm{Tr}(\boldsymbol{Q}_1^T \boldsymbol{X}_1^T \boldsymbol{X}_2 \boldsymbol{Q}_2)$ subject to the normalization constraints in (1b) is equivalent to maximizing the cross-correlation between the reduced-dimension views $\boldsymbol{G}_1 = \boldsymbol{X}_1 \boldsymbol{Q}_1$ and $\boldsymbol{G}_2 = \boldsymbol{X}_2 \boldsymbol{Q}_2$. When there are more than two views, extensions of (1) are usually referred to as *generalized canonical correlation analysis* (GCCA) (or, occasionally, *multi-set CCA*). GCCA can employ various objective functions. The most natural extension of the two-view CCA is arguably the so-called sum-of-correlations (SUMCOR) GCCA [32], [2], [33], [34], [3], [31], [19], which aims at solving the following problem:

$$\max_{\{\boldsymbol{Q}_i\}_{i=1}^I} \ \sum_{i=1}^I \sum_{j \neq i}^I \mathrm{Tr}\left(\boldsymbol{Q}_i^T \boldsymbol{X}_i^T \boldsymbol{X}_j \boldsymbol{Q}_j\right) \tag{2}$$

$$\text{s.t.} \ \boldsymbol{Q}_i^T \boldsymbol{X}_i^T \boldsymbol{X}_i \boldsymbol{Q}_i = \boldsymbol{I}_K, \ i=1,\ldots,I.$$

SUMCOR looks for common structure in multiple views via enforcing pairwise similarities between the reduced-dimension views, i.e., $\boldsymbol{X}_i \boldsymbol{Q}_i$'s. Unlike the two-view CCA whose optimal solution amounts to an eigen-decomposition, SUMCOR is NP-hard [19], [33]. In this work, our major interest lies in handling SUMCOR GCCA and its variants. Note that SUMCOR is not the only GCCA formulation. Many other formulations such as MAX-VAR and SUQUAR exist; see [3], [2].

## B. Structured GCCA

In many applications, plain GCCA/CCA is not enough to produce satisfactory results in terms of learning meaningful and interpretable latent representations. This is because there is always noise and model mismatches when dealing with real-world data. In those cases, effectively using any available prior information on the canonical components becomes crucial. This leads to a so-called structured GCCA formulation

$$\min_{\{Q_i\}_{i=1}^I} -\sum_{i=1}^I \sum_{j\neq i}^I \text{Tr}\left(Q_i^T X_i^T X_j Q_j\right) + \sum_{i=1}^I \lambda_i r_i(Q_i) \quad (3)$$
$$\text{s.t.} \quad Q_i^T X_i^T X_i Q_i = I_K, \; i=1,\ldots,I,$$

where the regularization terms reflect prior knowledge, interpreted as coming from a prior distribution in a Bayesian setting. Several structural regularizations are of particular interest. For example,

$$r_i(Q_i) = \|Q_i\|_1 \quad (4)$$

has been frequently used in the literature for feature selection. Zeros in $Q_i$ will null columns in $X_i$ when forming $X_i Q_i$. This can effectively suppress irrelevant features. It can also be used for identifying highly relevant features from multiple views – which is a very important application of GCCA in genetics [12], [13], [14]. Another very popular feature selection regularizer is

$$r_i(Q_i) = \|Q_i\|_{2,1}, \quad (5)$$

which promotes zero rows in $Q_i$. This would lead to a dimension-reducing matrix $Q_i$ that selects the same columns in $X_i$ to produce $X_i Q_i$. Fig. 1 gives an illustration of the effect of group-sparsity in $Q_i$: the zero rows of $Q_i$ exclude the irrelevant features (columns) in $X_i$ when producing the latent representations.

In addition, elastic net regularizers like

$$r_i(Q_i) = \|Q_i\|_1 + \|Q_i\|_F^2, \; r_i(Q_i) = \|Q_i\|_{2,1} + \|Q_i\|_F^2 \quad (6)$$

are also considered [21], since the Frobenius norm can help improve the numerical stability in some cases where badly conditioned subproblems are encountered during the optimization process. There are many other constraints that might be of interest to practitioners in different domains such as nonnegativity and smoothness of $Q_i$ [13]. In the current work, the proposed approach can handle all the aforementioned regularizers as well as any other regularizing function with a tractable proximal operator.

## C. Challenges and Prior Art

The SUMCOR GCCA problem in (2) is NP-hard [19], [33], but effective approximations exist. For example, the algorithm in [19] proposed to handle (2) as follows: first, let $Z_i = (X_i^T X_i)^{1/2} Q_i$. Then, Problem (2) can be written as

$$\min_{\{Z_i\}} \sum_{i=1}^I \sum_{j\neq i}^I \text{Tr}\left(Z_i^T (X_i^T X_i)^{-1/2} X_i^T X_j (X_j^T X_j)^{-1/2} Z_j\right) \quad (7a)$$
$$\text{s.t.} \quad Z_i^T Z_i = I_K. \quad (7b)$$

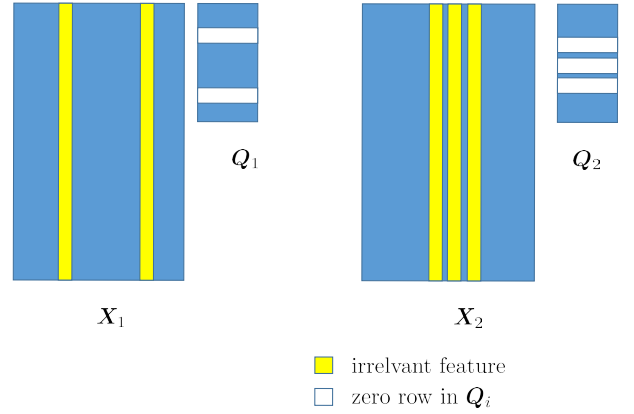

Fig. 1: Illustration of group-sparsity structured CCA.

Then, an alternating optimization algorithm is proposed to solve the above w.r.t. $Z_i$ when fixing $Z_j$'s – which amounts to SVD of

$$F_i = \sum_{j=1, j\neq i}^I (X_i^T X_i)^{-1/2} X_i^T X_j (X_j^T X_j)^{-1/2} Z_j.$$

The algorithm works fairly well as observed in [23] – but only for small-scale cases where both $L, M_i \leq 10,000$. The difficulty in computations lies in the term $(X_j^T X_j)^{-1/2}$, which is a *dense* $M_i \times M_i$ matrix even when $X_i$ is sparse – if $M_i = 10,000$, this matrix costs 75 GB memory when double precision is employed. In addition, computing the inverse of $(X_i X_i)^{1/2}$ consumes $\mathcal{O}(10^{12})$ flops, which is also too costly. When structured GCCA is considered, the work in [32] takes a similar approach using the same change of variables to handle the problem, which is, again, not scalable.

To overcome the issues in large-scale GCCA computations, Fu et al. [8] proposed another framework for alternating optimization. The idea is to make use of the following change of variables:

$$X_i Q_i = G_i \Leftrightarrow Q_i = (X_i^T X_i)^{-1} X_i^T G_i, \; G_i \in \mathcal{R}(X_i), \quad (8)$$

where $\mathcal{R}(\cdot)$ denotes the range space of a matrix and rewrite the SUMCOR GCCA problem as:

$$\min_{\{G_i\}} \sum_{i=1}^I \sum_{j\neq i}^I \text{Tr}\left(G_i^T X_i (X_i^T X_i)^{-1} X_i^T X_j (X_j^T X_j)^{-1} X_j^T G_j\right) \quad (9a)$$
$$\text{s.t.} \quad G_i^T G_i = I_K, \; G_i \in \mathcal{R}(X_i). \quad (9b)$$

Then, alternating optimization is applied to $G_i$ for $i = 1, \ldots, I$. Every block is solved by applying SVD to

$$H_i = X_i (X_i^T X_i)^{-1} X_i^T \sum_{j=1, j\neq i}^I X_j (X_j^T X_j)^{-1} X_j^T G_j.$$

Note that $H_i$ can be constructed much more easily compared to $F_i$: the most difficult part $(X_j^T X_j)^{-1} X_j^T G_j$ can be obtained by any iterative least squares (LS) algorithm such as conjugate gradient [35]. The total memory cost is $\mathcal{O}(LK)$ and the computational cost is in the order of $\mathcal{O}(\text{nnz}(X_i)K)$, where nnz stands for the number of non-zero elements of an array. The work in [8] successfully scaled SUMCOR up to handle views of size $100,000 \times 100,000$. However, the whole framework relies on (8), which only holds when the plain SUMCOR problem in (2) is considered. If one considers a structured SUMCOR problem with a regularization term $r_i(\boldsymbol{Q}_i)$, the framework in [8] cannot be applied. However, structured large-scale GCCA is very much desired in practice.

## III. PROPOSED ALGORITHM

In this section, we address the challenges for large-scale structured GCCA. Specifically, we will propose an algorithmic framework that is able to handle huge-scale multiview data under a variety of structural regularizations on the canonical components – with affordable memory and computational costs. The proposed framework can also easily facilitate parallel computations with limited communication overhead.

### A. First Attempt: ADMM

To begin with, let us consider the alternative formulation of structured GCCA:

$$\min_{\{\boldsymbol{Q}_i\}_{i=1}^I} \sum_{i=1}^I \sum_{j>i}^I \frac{1}{2} \|\boldsymbol{X}_i\boldsymbol{Q}_i - \boldsymbol{X}_j\boldsymbol{Q}_j\|_F^2 + \sum_{i=1}^I \lambda_i r_i(\boldsymbol{Q}_i) \quad (10)$$
$$\text{s.t. } \boldsymbol{Q}_i^T \boldsymbol{X}_i^T \boldsymbol{X}_i \boldsymbol{Q}_i = \boldsymbol{I}_K, \ i=1,\ldots,I.$$

Note that if one expands the first term in the objective and discard the constants, the formulation in (3) is recovered. To proceed, we introduce some slack variables to re-write the above as

$$\min_{\{\boldsymbol{Q}_i,\boldsymbol{G}_i\}_{i=1}^I} \sum_{i=1}^I \sum_{j>i}^I \frac{1}{2} \|\boldsymbol{X}_i\boldsymbol{Q}_i - \boldsymbol{G}_j\|_F^2 + \sum_{i=1}^I \lambda_i r_i(\boldsymbol{Q}_i) \quad (11a)$$
$$\text{s.t. } \boldsymbol{G}_i = \boldsymbol{X}_i\boldsymbol{Q}_i, \ \forall i, \quad (11b)$$
$$\boldsymbol{G}_i^T \boldsymbol{G}_i = \boldsymbol{I}_K, \ \forall i, \quad (11c)$$

where $\boldsymbol{G}_i \in \mathbb{R}^{L \times K}$ is a thin matrix. This way, we have not changed the optimization problem, but have managed to split the challenge into easier pieces using different variables, as will be seen shortly. To deal with Problem (11), we propose to employ a primal-dual approach. Specifically, we consider the augmented Lagrangian of Problem (11), which is

$$\mathcal{L}\left(\{\boldsymbol{Q}_i,\boldsymbol{G}_i,\boldsymbol{Y}_i\}_{i=1}^I\right)$$
$$= \sum_{i=1}^I \sum_{j>i}^I \frac{1}{2} \|\boldsymbol{X}_i\boldsymbol{Q}_i - \boldsymbol{G}_j\|_F^2 + \sum_{i=1}^I \lambda_i r_i(\boldsymbol{Q}_i) \quad (12)$$
$$+ \frac{\rho}{2} \sum_{i=1}^I \left\|\boldsymbol{X}_i\boldsymbol{Q}_i - \boldsymbol{G}_i + \frac{1}{\rho}\boldsymbol{Y}_i\right\|_F^2,$$

where $\boldsymbol{Y}_i$ is the dual variable associated with the equality constraint $\boldsymbol{X}_i\boldsymbol{Q}_i = \boldsymbol{G}_i$. To find $\boldsymbol{G}_i$ and $\boldsymbol{Q}_i$, a natural thought would be to invoke the *alternating direction method of multipliers* (ADMM), which handles $\mathcal{L}\left(\{\boldsymbol{Q}_i,\boldsymbol{G}_i,\boldsymbol{Y}_i\}_{i=1}^I\right)$ w.r.t. $\{\boldsymbol{Q}_i\}_i$, $\{\boldsymbol{G}_i\}_i$ and $\{\boldsymbol{Y}_i\}_i$ in a cyclical manner. Specifically, a standard ADMM algorithm updates the variables as follows

$$\boldsymbol{Q}_i \leftarrow \arg\min_{\boldsymbol{Q}_i} \sum_{j=1,j\neq i}^I \frac{1}{2} \|\boldsymbol{X}_i\boldsymbol{Q}_i - \boldsymbol{G}_j\|_F^2 + \lambda_i r_i(\boldsymbol{Q}_i)$$
$$+ \frac{\rho}{2} \left\|\boldsymbol{X}_i\boldsymbol{Q}_i - \boldsymbol{G}_i + \frac{1}{\rho}\boldsymbol{Y}_i\right\|_F^2 \quad (13a)$$

$$\boldsymbol{G}_i \leftarrow \arg\min_{\boldsymbol{G}_i^T\boldsymbol{G}_i=\boldsymbol{I}_K} \sum_{j=1,j\neq i}^I \frac{1}{2} \|\boldsymbol{X}_j\boldsymbol{Q}_j - \boldsymbol{G}_i\|_F^2$$
$$+ \frac{\rho}{2} \left\|\boldsymbol{X}_i\boldsymbol{Q}_i - \boldsymbol{G}_i + \frac{1}{\rho}\boldsymbol{Y}_i\right\|_F^2 \quad (13b)$$

$$\boldsymbol{Y}_i \leftarrow \boldsymbol{Y}_i + \rho(\boldsymbol{X}_i\boldsymbol{Q}_i - \boldsymbol{G}_i). \quad (13c)$$

Note that the updates in (13) are already a very appealing for large-scale GCCA. First, the subproblem w.r.t. $\boldsymbol{Q}_i$ is a regularized least squares problem, which can be solved using various lightweight algorithms such as the *fast iterative soft thresholding algorithm* (FISTA) [36] and *proximal gradient* (PG) [37]. Specifically, if PG is employed, then $\boldsymbol{Q}_i$ is updated by the following subroutine:

$$\boldsymbol{Q}_i \leftarrow \arg\min_{\boldsymbol{Q}_i} \left\|\boldsymbol{Q}_i - \left(\hat{\boldsymbol{Q}}_i - \alpha\nabla f(\hat{\boldsymbol{Q}}_i)\right)\right\|_F^2 + \lambda_i r_i(\boldsymbol{Q}_i), \quad (14)$$

where $\hat{\boldsymbol{Q}}_i$ is the previous iterate of $\boldsymbol{Q}_i$, and

$$\nabla f(\hat{\boldsymbol{Q}}_i) = \sum_{j=1,j\neq i}^I (\boldsymbol{X}_i^T\boldsymbol{X}_i\boldsymbol{Q}_i - \boldsymbol{X}_i^T\boldsymbol{G}_j)$$
$$+ \rho(\boldsymbol{X}_i^T\boldsymbol{X}_i\boldsymbol{Q}_i - \boldsymbol{X}_i^T(\boldsymbol{G}_i - 1/\rho\boldsymbol{Y}_i))$$

is the gradient of the smooth part of the cost function in (13a). One can see that, to implement PG, a complexity order of $\mathcal{O}(\text{nnz}(\boldsymbol{X}_i)K)$ flops per iteration is the cost to compute the gradient, since the multiplications such as $\boldsymbol{X}_i^T\boldsymbol{G}_j$, $\boldsymbol{X}_i\boldsymbol{Q}_i$ and $\boldsymbol{X}_i^T(\boldsymbol{X}_i\boldsymbol{Q}_i)$ all consume $\mathcal{O}(\text{nnz}(\boldsymbol{X}_i)K)$ flops. This is very lightweight when the views are sparse (recall that $K$ is the number of canonical components, which is usually very small compared to $L$ and $M_i$).

Second, the subproblem in (13b) can be solved by economy-size SVD. To be specific, an optimal solution to Problem (13b) is

$$\boldsymbol{U}_i\boldsymbol{\Sigma}_i\boldsymbol{V}_i^T \leftarrow \text{svd}\left(\sum_{j\neq i} \boldsymbol{X}_j\boldsymbol{Q}_j + \rho\boldsymbol{X}_i\boldsymbol{Q}_i + \boldsymbol{Y}_i, '\text{econ}'\right), \quad (15a)$$

$$\boldsymbol{G}_i \leftarrow \boldsymbol{U}_i\boldsymbol{V}_i^T, \quad (15b)$$

where $\boldsymbol{U}_i \in \mathbb{R}^{L \times K}$, $\boldsymbol{\Sigma}_i \in \mathbb{R}^{K \times K}$ and $\boldsymbol{V}_i \in \mathbb{R}^{K \times K}$ and svd$(\cdot, '\text{econ}')$ denotes the economy-size SVD, which only needs $\mathcal{O}(LK^2)$ flops to carry out.

Third, the update of $\boldsymbol{Y}_i$ also costs very few flops since $\boldsymbol{X}_i\boldsymbol{Q}_i$ has already been computed and only additions of thin matrices are left.

Note that solving all the subproblems cost $\mathcal{O}(LK)$ in terms of memory – which is very cheap. Finally, the ADMM structure is very friendly for parallel computing; i.e., $\boldsymbol{Q}_i$ and $\boldsymbol{G}_i$ for $i=1,\ldots,I$ can be updated simultaneously by different computing agents. What needs to be exchanged among the agents is the $G_i$ and $Y_i$, which are thin matrices of size $L \times K$, and thus do not cost much communication overhead.

The ADMM algorithm looks very promising in terms of accommodating large-scale views, but there is a caveat: the structured SUMCOR problem is a nonconvex optimization with a manifold constraint. Therefore, the standard ADMM does not necessarily converge. Note that the risk of divergence is not desired – especially in the large-scale case where every iteration may cost a considerable amount of effort. This raises a natural question: is there a way to keep the nice computational advantages of ADMM while ensuring convergence to a stationary or KKT point? We will show that, with delicate design and the employment of adaptive step size $\rho$, the answer is thankfully affirmative.

### B. Penalty-Dual Decomposition (PDD)

To fix the issues with ADMM, we propose to employ the so-called *penalty-dual decomposition* (PDD) framework [29], [38] for solving the structured SUMCOR problem in (12). The PDD consists of similar updates as those in ADMM, but with different ordering and an adaptive $\rho$. This way, the per-iteration complexity of ADMM is not increased and most advantages are kept, while the algorithm is guaranteed to converge to the set of KKT points.

The PDD framework is presented in Algorithm (1). The PDD algorithm consists of two parts. The first part is a sub-solver that handles the augmented Lagrangian w.r.t. $\{Q_i\}$ and $\{G_i\}$ when $Y_i$'s are fixed. The parameter $\epsilon^{(r)}$ is for specifying the accuracy of the solution at iteration $r$ – and the definition of accuracy will be specified shortly.

The `sub-solver` is a solver for the following problem:

$$\min_{\{Q_i, G_i\}} \sum_{i=1}^{I}\sum_{j>i}^{I} \frac{1}{2}\|X_i Q_i - G_j\|_F^2 + \sum_{i=1}^{I} \lambda_i r_i(Q_i)$$
$$+ \frac{\rho^{(r)}}{2} \sum_{i=1}^{I} \left\| X_i Q_i - G_i + \frac{1}{\rho^{(r)}} Y_i^{(r)} \right\|_F^2 \quad (16)$$
$$\text{s.t. } G_i^T G_i = I_K, \ \forall i.$$

Note that the solver need not solve the above to optimality. Every call of the `sub-solver` only requires that $Q_i$ and $G_i$ converge to a neighborhood of a KKT point of (16), roughly speaking, where the 'diameter' of the neighborhood is specified by $\epsilon^{(r)}$.

One easily implementable solver for (16) is the so-called inexact alternating optimization. Specifically, one may employ the following updates alternately between $\{Q_i\}$ and $\{G_i\}$:

$$Q_i^+ \leftarrow \arg\min_{Q_i} \ \left\| Q_i - \left(\hat{Q}_i - \alpha \nabla f(\hat{Q}_i)\right) \right\|_F^2 + \lambda_i r_i(Q_i), \quad (17a)$$

$$G_i^+ \leftarrow \arg\min_{G_i^T G_i = I_K} \sum_{j=1, j\neq i}^{J} \frac{1}{2} \left\| X_j Q_j^+ - G_i \right\|_F^2$$
$$+ \frac{\rho^{(r)}}{2} \left\| X_i Q_i^+ - G_i + \frac{1}{\rho^{(r)}} Y_i \right\|_F^2 \quad (17b)$$

$$\hat{Q}_i \leftarrow Q_i^+, \ \hat{G}_i \leftarrow G_i^+. \quad (17c)$$

---

**Algorithm 1:** `PDD-GCCA`

**input :** $\{X_i\}_{i=1}^I$; $K$; $\rho^{(0)} > 0$; $0 < c < 1$; $\{\epsilon^{(r)}, \eta^{(r)}\}_{r=1}^\infty$.
1   $r \leftarrow 0$;
2   **repeat**
3      $\left(\{Q_i^{(r+1)}, G_i^{(r+1)}\}_{i=1}^I\right) \leftarrow$
       `sub-solver`$\left(\{Y_i^{(r)}\}_i, \rho^{(r)}, \epsilon^{(r)}\right)$ for (16);
4      **if** $\sum_{i=1}^{I} \|X_i Q_i^{(r+1)} - G_i^{(r+1)}\|_F^2 \leq \eta^{(r)}$ **then**
5          $Y_i^{(r+1)} = Y_i^{(r)} + \rho^{(r)}(XQ_i^{(r+1)} - G_i^{(r+1)})$,
         $\rho^{(r+1)} = \rho^{(r)}$;
6      **else**
7          $Y_i^{(r+1)} = Y_i^{(r)}$, $\rho^{(r+1)} = c\rho^{(r)}$;
8      **end**
9      $r \leftarrow r+1$;
10 **until** *some stopping criterion is reached*;
**output:** $\{G_i\}$

---

where $\hat{Q}_i$ and $Q_i^+$ are the old and new iterates of $Q_i$, and

$$\nabla f(\hat{Q}_i) = \sum_{j=1, j\neq i}^{I} (X_i^T X_i \hat{Q}_i - X_i^T \hat{G}_j)$$
$$+ \rho^{(r)}(X_i^T X_i \hat{Q}_i - X_i^T(\hat{G}_i - 1/\rho^{(r)} Y_i))$$

is the gradient of the smooth part of the objective as defined before. This `sub-solver` is summarized in Algorithm 2.

One may have already noticed that the PDD algorithm utilizes exactly the same 'ingredients' of ADMM – i.e., if one breaks down the two algorithms into subproblems, PDD and ADMM update $G_i$, $Q_i$ and $Y_i$ via solving exactly the same subproblems. This means that PDD inherits all the low-complexity advantages of ADMM. In fact, PDD can be considered a variant of ADMM, which also aims at solving the augmented Lagrangian of the primal problem while changing the weight of the penalty, i.e., $\rho$, along the iterates. This way, and along with a delicately designed updating order of the primal and dual variables, convergence of the algorithm can be guaranteed.

We should mention that the update of $Q_i$ in (17a) given a fixed $G_i$ may be performed more than once. One can carry out the update of $Q_i$ multiple times and then update $G_i$. This can effectively reduce the total number of alternating iterations between $Q_i$ and $G_i$. This may not necessarily reduce the total amount of computations, but is very meaningful when a distributed algorithm is considered. The reason is that when the algorithm shifts from updating $Q_i$ to $G_i$, all the information about $X_i Q_i$ needs to be shared among the agents, which incurs communication overhead. Although the $X_i Q_i$'s are thin matrices which are relatively cheap to exchange, it is still good to reduce overhead via reducing the frequency of alternating between the $Q_i$-subproblem and $G_i$-subproblem. Another observation is that if one updates $Q_i$ multiple times until convergence of $Q_i$ and then updates $G_i$ without going back to $Q_i$ again (and then updates $Y_i$), we recover the ADMM updates if $\rho$ is also fixed.

The PDD framework opens the door to establishing convergence guarantees, but proving convergence for a particular application of PDD, including to GCCA, is generally far from trivial. In the next section, we present detailed convergence analysis of `PDD-GCCA`.

**Algorithm 2:** sub-solver

**input**: $\{Y^{(r)}, Q_i^{(r)}, G_i^{(r)}\}_i$; $\epsilon^{(r)}$.
1. $\hat{G}_i \leftarrow G_i^{(r)}, \hat{Q}_i \leftarrow Q_i^{(r)}$;
2. **repeat**
3. $\quad \nabla f(\hat{Q}_i) \leftarrow \sum_{j=1,j\neq i}^{I}(X_i^T X_i \hat{Q}_i - X_i^T \hat{G}_j) + \rho^{(r)}(X_i^T X_i \hat{Q}_i - X_i^T(\hat{G}_i - 1/\rho^{(r)} Y_i^{(r)}))$;
4. $\quad Q_i^+ \leftarrow \arg\min_{Q_i} \left\| Q_i - \left(\hat{Q}_i - \alpha \nabla f(\hat{Q}_i)\right)\right\|_F^2 + \lambda_i r_i(Q_i)$;
5. $\quad U_i \Sigma_i V_i^T \leftarrow$ svd $\left(\sum_{j\neq i} X_j Q_j^+ + \rho^{(r)} X_i Q_i^+ + Y_i^{(r)}, '\text{econ}'\right)$,
6. $\quad G_i^+ \leftarrow U_i V_i^T$;
7. $\quad \hat{Q}_i \leftarrow Q_i^+, \hat{G}_i \leftarrow G_i^+$;
8. **until** $\max\left(\|\hat{Q}_i - Q_i^+\|_\infty, \|\hat{G}_i - G_i^+\|_\infty\right) \leq \epsilon^{(r)}$;
9. $G_i^{(r+1)} \leftarrow G_i^{(r)}, Q_i^{(r+1)} \leftarrow Q_i^{(r)}$;
**output**: $\{G_i^{(r+1)}, Q_i^{(r+1)}\}$

## IV. CONVERGENCE ANALYSIS

On a high-level, the PDD algorithm can be considered as a judicious combination of the penalty method and the augmented Lagrangian method in optimization – which are both long-existing techniques for handling constrained optimization problems. The penalty method 'lifts' the equality constraints as a fitting penalty term to the objective function, and gradually increases the penalty parameter to infinity, thereby enforcing feasibility of the original problem. The penalty method works in theory even when the problem is nastily nonconvex. The issue is that when the penalty parameter is too large, the corresponding subproblem is extremely ill-conditioned and the numerical stability in practice is severely hindered. Lagrangian methods introduce a dual variable to help expedite enforcing feasibility, which avoids the ill-conditioning problem in the penalty method effectively – but they usually do not guarantee convergence when there is nonconvexity involved.

PDD can be considered somewhere in between. It employs dual variables associated with the relaxed equality constraints and it also gradually increases the penalty parameter $\rho$ – but in a much more controllable way: when the relaxed constrained is satisfied to some extent, PDD does not increase $\rho$ (cf. line 4-5 in Algorithm 1); when the constraints are largely violated, then a larger $\rho$ is used to put more emphasis on the augmented penalty term (line 7). This way, PDD can have the ability of handling nonconvex problems as the penalty method does, but also use help from the Lagrangian dual method to help the feasibility part, thereby decreasing the speed of increasing $\rho$ – hopefully, the algorithm will achieve a good enough solution before $\rho$ gets too large.

To formally prove convergence of PDD-based GCCA, we need to check if the algorithm design and the problem structure satisfy the convergence requirements of the PDD framework. To this end, we first show the following intermediate lemma.

**Lemma 1** *The solution sequence produced by the* sub-solver *in Algorithm 2 converges to the set of KKT points of Problem* (16).

*Proof:* There proof consists of two steps. First, the alternating optimization algorithm between $\{G_i\}$ and $\{Q_i\}$ is special in the sense that the $G_i$'s are constrained on orthogonal manifolds which are nonconvex. Therefore, the classical results for alternating optimization, such as those in [39], [40] cannot be directly applied. However, as shown in [41], [7], the two-block structure guarantees that every limit point of the produced solution sequence is a KKT point of Problem (16). Second, following the arguments in [41], [7], one can further show that all the produced $\{Q_i\}$ and $\{G_i\}$ are bounded – therefore, the whole sequence converges to a KKT point of Problem (16). ∎

Lemma 1 is important for convergence of the overall PDD-GCCA algorithm. It also ensures that the stopping criterion in line 8 of Algorithm 2 can be reached using the proposed sub-solver. Next, as another stepping stone to showing convergence of PDD-GCCA, we show that:

**Proposition 1** *Assume that* $\min\left(L, \frac{1}{I}\sum_{i=1}^{I} M_i\right) \geq \frac{K+1}{2}$. *Then any set of points* $\{Q_i, G_i\}_{i=1}^{I}$ *generated by Algorithm 1 satisfies Robinson's condition.*

The details of Robinson's condition can be found in the Appendix. This condition is a regularity condition that is crucial for showing convergence of a PDD algorithm. Note that Robinson's condition is non-trivial, and verifying it is non-trivial either – as can be appreciated in the Appendix. According to Proposition 1, Robinson's condition easily holds in practice, since we always have $K \ll \min\{M_i, L\}$ – i.e., the number of canonical components is much smaller than the data dimensions.

With Lemma 1 and Proposition 1, we can easily show the following proposition by invoking Theorem 3.1 in [38]:

**Proposition 2** *Assume that* $\epsilon^{(r)} \to 0$, $\eta^{(r)} \to 0$ *as* $r \to \infty$ *and that the stopping criterion of the* sub-solver *involved in Algorithm 1 is*

$$\max\left(\|\hat{Q}_i - Q_i^+\|_\infty, \|\hat{G}_i - G_i^+\|_\infty\right) \leq \epsilon^{(r)}, \quad \forall r. \quad (18)$$

*Then, every limit point of the solution sequence produced by the proposed* PDD-GCCA *algorithm is a KKT point of Problem* (10).

*Proof:* With Lemma 1 and Proposition 1 in our hands, the proof is a direct application of the result in [38] and thus is omitted. ∎

## V. NUMERICAL RESULTS AND EXPERIMENTS

In this section we demonstrate the performance of the proposed algorithm PDD-GCCA and showcase it's effectiveness in synthetic- and real-data experiments. All simulations are implemented in Matlab and are executed on a Linux server comprising 32 cores at 2GHz and 128GB RAM.

### A. Synthetic-Data Experiments

In this subsection we first showcase the effectiveness of PDD-GCCA using large-scale simulations. The multiple views are generated as follows. We assume that the views share a common latent factor $S \in \mathbb{R}^{L \times M_i}$ which is a randomly generated sparse matrix whose non-zero entries follow the zero-mean unit-variance Gaussian distribution. Each view $\boldsymbol{X}_i$ is generated following $\boldsymbol{X}_i = \boldsymbol{S}\boldsymbol{A}_i$, where $\boldsymbol{A}_i \in \mathbb{R}^{M_i \times M_i}$ is a matrix that maps the shared factor to the $i$th view. The density level of $\boldsymbol{S}$ and $\boldsymbol{A}_i$ is controlled such that the density level of each view, i.e., `density`$_i$ = $\frac{\text{nnz}(\boldsymbol{X}_i)}{LM_i}$, is controlled. The number of features of each view is set to be equal for simplicity, i.e. $M_i = M$.

We first test the algorithms when the views do not have outlying features. Under this setting, we let $r_i(\boldsymbol{Q}_i) = 0$ and (10) recovers the classic SUMCOR GCCA formulation; and we adopt the recently proposed large-scale SUMCOR algorithms `LasCCA` and `DisCCA` [8] as baselines. `LasCCA` and `DisCCA` were shown to be state-of-the-art when handling very large-scale and sparse mutiview data [8].

To evaluate the performance, we observe the total correlation captured. Since the views share a common latent factor, they are perfectly correlated in the shared latent domain. Therefore, the optimal value of total correlation is achieved when $\boldsymbol{A}_i\boldsymbol{Q}_i$'s are perfectly aligned with each other – which yield a SUMCOR value of $KI(I-1)$. We also record the time each algorithm needs to capture 95% of the optimal correlation, denoted as `95%-time`. The number of observations is $L = 120,000$ and each view has $M = 100,000$ features. The number of canonical components is set to $K = 5$. The parameter $\rho^{(0)}$ for `PDD-GCCA` is set to be 2. The maximal number of iterations of the sub-solver is set to be 5. The sequence for the primal residual is chosen to be $\eta^{(r)} = \frac{100}{r}$ and $c = 0.9$. The results are averaged over 20 Monte Carlo trials.

Table I shows the performance of the algorithms under various density levels. One can see that `LasCCA` works very well in terms of capturing correlations. Under all the tested density levels, `LasCCA` captures more than 98% of the total SUMCOR (which is 100 in this simulation). `DisCCA` also works well when `density` = $5 \times 10^{-5}$ and $10^{-4}$, but less competitive when `density` = $10^{-5}$. `PDD-GCCA` has comparable performance relative to `LasCCA` in terms of capturing correlations, but `PDD-GCCA` works much faster – when using a single core implementation, it is already at least 3 times faster than the benchmarking algorithms. This may be because `PDD-GCCA` uses a primal-dual framework, and dual updates usually help expedite convergence – while both `LasCCA` and `DisCCA` are primal algorithms which do not employ any dual updates. In addition, if one employs a multicore implementation of `PDD-GCCA` (recall that `PDD-GCCA` can be implemented distributively using $I$ cores), the runtime performance of `PDD-GCCA` is even better – as a result, the multicore version is at around 10 times faster than the baselines for capturing 95% of the total correlations. Fig. 2 shows the convergence curves of the algorithms over time. One can see that the `PDD-GCCA` algorithm converges faster than `LasCCA` and `DisCCA`. The multicore version of `PDD-GCCA` performs the best in terms of convergence time.

We also test the algorithms when outlying features are present in the views. To simulate such scenarios the data are generated as follows: $\boldsymbol{X}_i = [\boldsymbol{S}\boldsymbol{A}_i, \boldsymbol{O}_i] + \boldsymbol{N}$, where $\boldsymbol{S}$ and $\boldsymbol{A}_i$ are defined as before, $\boldsymbol{N}$ is zero mean 0.01 variance sparse Gaussian noise and $\boldsymbol{O}_i \in \mathbb{R}^{L \times M_o}$ is also a sparse matrix with zero-mean and unit variance non zero entries – but completely uncorrelated across the views. $M_o$ is the number of outlying features and $[\cdot,\cdot]$ denotes the concatenating operation of two matrices. The 'signal part' and the 'outlier' part of each view are forced to have comparable energy levels, i.e. $\|\boldsymbol{S}\boldsymbol{A}_i\|_F \approxeq \|\boldsymbol{O}_i\|_F$, so that simple energy detection would not identify the outlying features. In this case, GCCA has two objectives. The first is to capture the highest possible correlation between the informative part of the views, while at the same time to suppress the impact of $\boldsymbol{O}_i$. Towards this end, the $\ell_{2,1}$ and $\ell_1$ norms that promote (row-)sparsity are employed to serve as $r_i(\boldsymbol{Q}_i)$. In order to evaluate the performance of the algorithms, two metrics are introduced as in [7]. Let $\mathcal{I}_s$ and $\mathcal{I}_o$ be the index sets of the signal and the outlying columns in $\boldsymbol{X}_i$, respectively, where $\mathcal{I}_o \bigcup \mathcal{I}_s = \{1, \ldots, L\}$. The first metric is: `metric_1`= $\sum_{i=1}^{I}\sum_{j>i}^{I} \text{Tr}\left(\boldsymbol{Q}_i(\mathcal{I}_s,:)^T \boldsymbol{X}_i(:,\mathcal{I}_s)^T \boldsymbol{X}_j(:,\mathcal{I}_s)\boldsymbol{Q}_j(\mathcal{I}_s,:)\right) \frac{100}{KI(I-1)}$, which measures the percentage of total signal correlation captured. In our simulations, `metric_1` $\in [0, 100]$, and a higher value of `metric_1` is desired. The second metric measures the ability of identifying and suppressing the outlying part. To this end, we define `metric_2`= $\sum_{i=1}^{I}\|\boldsymbol{Q}_i(\mathcal{I}_o,:)\|_F$ whose optimal value is zero and smaller values of `metric_2` correspond to better performance in suppressing outliers.

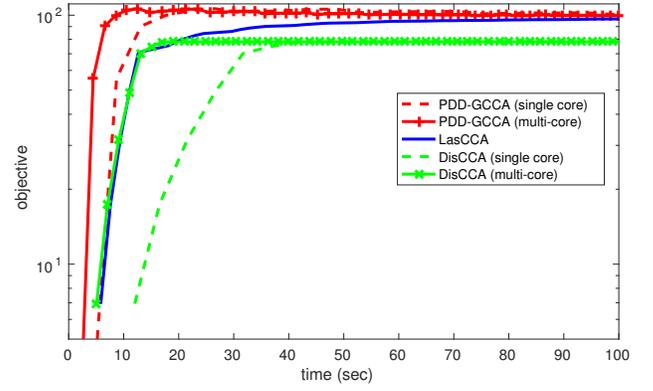

Fig. 2: Captured correlation over time of different algorithms.

Fig. 3 shows an illustrative example when outlying features are present. In this example, we have $L = 12,000$ and $M = M_o = 12,000$. In this case, we aim at extracting $K = 5$ canonical components. We apply `PDD-GCCA` with $\ell_{2,1}$ and $\ell_1$ norm regularizations and the baselines to this case and plot the aggregated row norms of $\boldsymbol{Q}_i$ for $i = 1, \ldots, I$. It is clear that the regularized GCCA algorithms can effectively suppress the outlying part.

Table II shows the performance of the algorithms using larger views. Here $L = 100,000$, and we have $80,000$ informative and $80,000$ outlying features. The number of canonical components varies from $K = 5$ to $K = 50$. The density level of each view is `density` = $10^{-4}$. The regularization parameter $\lambda_i$ is chosen to be 0.1. The results are averaged over 20 Monte Carlo simulations as before. One can see that `PDD-GCCA` with the $\ell_{2,1}$ norm regularization remarkably outperforms the unregularized ones: it captures more than 92% of the total signal correlation in all cases and it successfully suppresses the outlying features according to the values of `metric_2`. `PDD-GCCA` with $\ell_1$ regularizers also works well. This simulation also justifies our motivation of considering regularized GCCA: When outlying features are present, classic GCCA may not be able to produce satisfactory results.

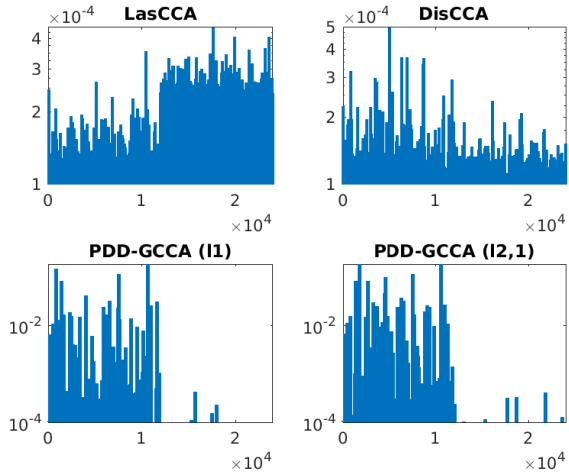

Fig. 3: squared row-sum vs rows

TABLE I: Evaluation of the algorithms; `PDD-GCCA` uses $r_i(\cdot) = 0$; $L = 120,000$ and $M = 100,000$.

| Algorithm | metric | density level | | |
|---|---|---|---|---|
| | | $10^{-4}$ | $5 \times 10^{-5}$ | $10^{-5}$ |
| `PDD-GCCA` (multicore) | corr. captured | 99.66 | **99.59** | 99.79 |
| | 95% time (sec) | **6.82** | **6.08** | **7.43** |
| `PDD-GCCA` | corr. captured | **99.67** | **99.59** | 99.79 |
| | 95% time (sec) | 22.06 | 14.64 | 18.57 |
| `LasCCA` | corr. captured | 98.96 | 98.77 | 99.37 |
| | 95% time (sec) | 59.56 | 63.23 | 87.5 |
| `DisCCA` (multicore) | corr. captured | 96.78 | 95.61 | 78.62 |
| | 95% time (sec) | 54.24 | 71.37 | inf |
| `DisCCA` | corr. captured | 96.78 | 95.61 | 78.62 |
| | 95% time (sec) | 133.20 | 144.97 | inf |

TABLE II: Performance of the algorithms in the presence of outliers.

| Algorithm | metric | # of canonical components | | | |
|---|---|---|---|---|---|
| | | $K=5$ | $K=10$ | $K=20$ | $K=50$ |
| `PDD-GCCA` | metric 1 | 27.78 | 28.44 | 29.21 | 29.95 |
| | metric 2 | 2.99 | 4.18 | 5.81 | 9.01 |
| `PDD-GCCA` ($\ell_1$) | metric 1 | 91.14 | **92.57** | 91.74 | 92.07 |
| | metric 2 | **0.55** | **0.88** | 1.35 | 2.34 |
| `PDD-GCCA` ($\ell_{2,1}$) | metric 1 | **92.26** | 92.43 | **94.72** | **95.48** |
| | metric 2 | 0.63 | 1.03 | **1.21** | **2.09** |
| `LasCCA` | metric 1 | 37.61 | 40.88 | 39.59 | 39.38 |
| | metric 2 | 2.51 | 3.41 | 4.84 | 7.55 |
| `DisCCA` | metric 1 | 18.28 | 18.44 | 17.61 | 17.31 |
| | metric 2 | 3.49 | 4.93 | 7.11 | 11.16 |

### B. Cross Language information Retrieval

In this subsection we apply the proposed algorithm to a cross-language information retrieval task to showcase the effectiveness in real-world applications. In this task, very high-dimensional representations of sentences in different languages are given. The objective is to learn latent low-dimensional representations of the sentences, which better reveal the correlations of the same sentences in different languages. This task is meaningful for query and search across languages since searching in high-dimensional space is very costly. It also helps machine translation [42], [43], [44].

**Data**: The data employed is the Europarl parallel corpus [45] (Europarl for short). The Europarl corpus is a collection of documents extracted from the proceedings of the European Parliament. It includes translated documents in 21 European languages: Romanic (French, Italian, Spanish, Portuguese, Romanian), Germanic (English, Dutch, German, Danish, Swedish), Slavik (Bulgarian, Czech, Polish, Slovak, Slovene), Finni-Ugric (Finnish, Hungarian, Estonian), Baltic (Latvian, Lithuanian), and Greek. Europarl provides multilingual alignment in the level of sentences. Each language presents a view, $\boldsymbol{X}_i$, and $\boldsymbol{X}_i(\ell, :)$ corresponds to the $\ell$th sentence in the $i$th language. Each sentence is represented by a high dimensional bag of words vector composed with inner-product preserving hashing [46], [47]. The number of hash slots used is $2^{19}$ which corresponds to $M = 2^{19} = 524,288$ features for each language. Seventy percent of the aligned sentences are randomly chosen as the training set, whereas twenty percent is used for testing. The remaining ten percent of the data was left in case validation is needed. The results in this paper did not use any validation during the training. The exact number of sentences used for training and testing depends on the languages used, since only a subset of sentences is aligned at all languages. Further details will be given in each experiment individually.

**Procedure**: We apply the GCCA algorithms to the training data to learn $\boldsymbol{Q}_i$ for $i = 1, \ldots, I$. At the testing stage, let $\hat{\boldsymbol{X}}_i$ be the testing set of the $i$-th language. For each query sentence $\hat{\boldsymbol{X}}_i(\ell, :)$ in the $i$-th language, Greek for example, we retrieve lower dimensional representation of $\hat{\boldsymbol{X}}_i(\ell, :)$ by computing $\hat{\boldsymbol{X}}_i(\ell, :)\boldsymbol{Q}_i$. Then all testing sentences of languages other than Greek are projected onto the associated lower dimensional space by computing $\hat{\boldsymbol{X}}_j \boldsymbol{Q}_j$ for $j \neq i$. Cross-language sentence retrieval is performed by selecting those testing sentences whose projections are closest to the low dimensional query in each language.

**Evaluation**: To measure the performance quantitatively, we define the distance:

$$\text{dist}_j^i(\ell, m) = \|\hat{\boldsymbol{X}}_i(\ell, :)\boldsymbol{Q}_i - \hat{\boldsymbol{X}}_j(m, :)\boldsymbol{Q}_j\|_2, \ m \in \mathcal{T},$$

where $\mathcal{T}$ is the index set of the testing sentences. Since we know that $\hat{\boldsymbol{X}}_i(\ell, :)$ and $\hat{\boldsymbol{X}}_j(\ell, :)$ correspond to the same sentence, we wish $\text{dist}_j^i(\ell, \ell)$ to be the smallest among all $\text{dist}_j^i(\ell, m)$'s for all $m$. Hence, for a given query language $i$, we denote $p_j^i(\ell)$ the position of $\text{dist}_j^i(\ell, \ell)$ after sorting $\text{dist}_j^i(\ell, m)$ for $m \in \mathcal{T}$ in an ascending order. If $p_j^i(\ell) = 1$ or a small integer, it means that the learned latent space reveals the correlation between the $\ell$th corresponding sentences well. If $p_j^i(\ell)$ is large, it means that there is some undesired distortion. The precision of sentence retrieval for language $j$ is measured by using the average Area under the Receiver Operating Characteristic curve (AROC) [48], [49], i.e.

$$\text{AROC} = \left(1 - \frac{p_j^i(\ell) - 1}{|\mathcal{T}| - 1}\right) \times 100\%,$$

with AROC$= 100\%$ being the best accuracy performance. The performance is measured via averaging the AROCs over all languages.

We also employ an extra metric to assess the use of the proposed algorithm in this specific application. That is, we count the frequency of $\hat{\boldsymbol{X}}_j(\ell,:)$ being the closest nearest neighbor (NN) of $\hat{\boldsymbol{X}}_i(\ell,:)$. Specifically, given a query language $i$ and sentence $\ell$, if $p_j^i(\ell) = 1$ then let $w_j^i(\ell) = 1$, otherwise $w_j^i(\ell) = 0$. Then, we define

$$\text{NN\_freq}(i,j) = \left(\frac{1}{|\mathcal{T}|} \sum_{\ell \in \mathcal{T}} w_j^i(\ell)\right) \times 100\%.$$

for language $i$ and $j$. We present the results averaged over all pairs of languages. Unlike the first metric that gives credit to $p_j^i(\ell) > 1$, the second metric only takes into account the cases where the latent representations of sentence $\ell$ in different languages being exactly the most correlated ones.

**Results**: Table III shows the performance of proposed algorithms when joint sentence retrieval is attempted among 6 languages, English, Greek, French, Italian, German and Spanish. The number of training sentences is $L = 430,816$, while the number of testing ones is $L_t = 61,545$. The vocabulary size is $M = 524,288$ words. The number of canonical components varies from 5 to 100. For `PDD-GCCA` with $\ell_{2,1}$ and $\ell_1$ regularizations, the penalty parameters $\rho$, $\rho_1$ are set to be equal to 2 and $\lambda$ is chosen to be equal to $10^{-3}$. The termination criterion is the total number of iterations reaching 20. All the algorithms are initialized using the lightweight multiview analysis algorithm `MVLSA` in [9]. One can see that the proposed algorithms show the best performance under the two considered metrics. All the algorithms perform better when $K$ increases. In particular, the proposed algorithms output much better nearest neighbor frequencies compared to the baselines.

TABLE III: Average AROC and NN freq. (both in %) of the English-Greek-French-Italian-Spanish-German joint cross language sentence retrieval experiment.

| Algorithm | metric | # of canonical components | | |
|---|---|---|---|---|
| | | $K = 5$ | $K = 70$ | $K = 100$ |
| `PDD-GCCA` | AROC | **96.89** | 97.63 | 97.60 |
| | NN freq. | **10.85** | 59.47 | 61.55 |
| `PDD-GCCA` ($\ell_1$) | AROC | 96.56 | **97.90** | 98.14 |
| | NN freq. | 8.99 | 57.71 | 61.13 |
| `PDD-GCCA` ($\ell_{2,1}$) | AROC | 96.56 | **97.90** | **98.15** |
| | NN freq. | 9.00 | 57.73 | 61.15 |
| `LasCCA` | AROC | 95.70 | 96.61 | 96.55 |
| | NN freq. | 6.97 | 48.88 | 50.31 |
| `DisCCA` | AROC | 93.05 | 96.61 | 97.42 |
| | NN freq. | 2.69 | 40.70 | 50.08 |
| `MVLSA` | AROC | 94.20 | 95.17 | 95.02 |
| | NN freq. | 3.82 | 30.63 | 31.11 |

Table IV shows the performance of the proposed algorithms when joint sentence retrieval is attempted among all 21 aligned languages. In this case the number of training sentences is $L = 134,228$, while the number of testing ones is equal to $L_t = 19,175$. The vocabulary size is $M = 524,288$ words. One can see that the proposed algorithms substantially outperform the baselines. In this case, the effect of the regularizers is more obvious especially in terms of finding the right nearest neighbor across languages. In particular, when $K = 100$ `PDD-GCCA` with the $\ell_{2,1}$ regularizer finds the right nearest neighbor 72 out of 100 times whereas the unregularized version of `PDD-GCCA` has a 5 percent lower NN freq. performance. Part of the reason why regularizers help is as follows. When we consider 21 languages, the number of the aligned sentences is smaller than before but the vocabulary remains the same. As a result, the views are represented as 'fat' matrices and thus the classical SUMCOR without regularization is not a well-defined formulation. Another observation is that, compared to the 6 view case, using 21 aligned views improves the retrieval performance since more information is added and processed. When $K = 100$ the achieved retrieval accuracy is 99.36%, meaning that we only need to scan the closest 0.64% of the sentences to collect the exact translation.

TABLE IV: Average AROC and NN freq. (both in %) of all 21 language joint cross sentence retrieval experiment.

| Algorithm | metric | # of canonical components | | |
|---|---|---|---|---|
| | | $K = 5$ | $K = 70$ | $K = 100$ |
| `PDD-GCCA` | AROC | **97.31** | 98.62 | 98.56 |
| | NN freq. | **7.49** | 65.88 | 67.48 |
| `PDD-GCCA` ($\ell_1$) | AROC | 97.20 | **99.27** | **99.36** |
| | NN freq. | 7.30 | 68.75 | 71.70 |
| `PDD-GCCA` ($\ell_{2,1}$) | AROC | 97.21 | **99.27** | **99.36** |
| | NN freq. | 7.31 | **68.77** | **71.72** |
| `LasCCA` | AROC | 95.77 | 97.55 | 97.61 |
| | NN freq. | 4.54 | 52.90 | 55.32 |
| `DisCCA` | AROC | 93.39 | 95.20 | 94.35 |
| | NN freq. | 2.20 | 27.18 | 26.54 |
| `MVLSA` | AROC | 93.67 | 95.19 | 94.37 |
| | NN freq. | 2.35 | 26.94 | 26.29 |

The results in Tables III and IV are obtained via testing on multiple languages. Another setup that is of interest in practice is to evaluate the result on two languages. Specifically, it would be interesting to see if learning $\boldsymbol{Q}_i$ from multiple languages could benefit translating between two particular languages. Note that the expectation of performance enhancement compared to only training with these two languages makes sense – since incorporating more languages (and thus richer information and more data samples) into the training process could benefit the testing stage, even though we are not testing on most of the languages.

The first set of experiments considers the case where the testing is performed between Greek and German. In this case the number of training sentences is $L = 134,228$, while the number of testing sentences is equal to $L_t = 19,175$. The vocabulary size is $M = 524,288$ words. The number of canonical components is set to $K = 70$ and the algorithms are run for 20 iterations. One can see that incorporating multiple languages gives clear benefits – both metrics show performance improvement along with increasing the number of languages involved in the training stage. When the number of languages is large, the regularized versions of `PDD-GCCA` again exhibit the best performance. Similar results can be seen in Table VI, where the same evaluation is performed on Greek and Finish. These two sets of experiments clearly show that using the proposed `PDD-GCCA` algorithm with regularization can help enhance cross-language sentence retrieval between two languages by training the latent representations from more languages.

TABLE V: Average AROC and NN freq. (both in %) of the German-Greek cross language sentence retrieval experiment.

| Algorithm | metric | # of languages | | | | |
|---|---|---|---|---|---|---|
| | | 2 | 3 | 5 | 10 | 21 |
| `PDD-GCCA` | AROC | **98.61** | **98.60** | 98.34 | 98.41 | 98.45 |
| | NN freq. | **63.65** | **65.16** | 63.85 | 65.27 | 65.40 |
| `PDD-GCCA` ($\ell_1$) | AROC | 98.35 | 98.55 | **98.72** | 99.08 | **99.18** |
| | NN freq. | 57.89 | 60.58 | 63.33 | 67.74 | 68.92 |
| `PDD-GCCA` ($\ell_{2,1}$) | AROC | 98.36 | 98.56 | **98.72** | **99.09** | **99.18** |
| | NN freq. | 58.01 | 60.66 | 63.38 | **67.76** | **68.93** |
| `LasCCA` | AROC | 97.14 | 97.27 | 97.13 | 97.30 | 97.35 |
| | NN freq. | 52.49 | 52.92 | 50.54 | 50.94 | 51.66 |
| `DisCCA` | AROC | 97.88 | 97.83 | 97.76 | 96.41 | 96.04 |
| | NN freq. | 57.02 | 50.04 | 49.49 | 35.82 | 29.76 |
| `MVLSA` | AROC | 96.02 | 95.94 | 95.18 | 94.70 | 96.04 |
| | NN freq. | 33.68 | 40.57 | 28.61 | 27.22 | 29.77 |

TABLE VI: Average AROC and NN freq. (both in %) of the Finish-Greek cross language sentence retrieval experiment.

| Algorithm | metric | # of languages | | | | |
|---|---|---|---|---|---|---|
| | | 2 | 3 | 4 | 10 | 21 |
| `PDD-GCCA` | AROC | **98.21** | 98.45 | 98.44 | 98.03 | 98.12 |
| | NN freq. | **57.09** | 59.99 | **61.15** | 56.95 | 57.99 |
| `PDD-GCCA` ($\ell_1$) | AROC | 98.13 | 98.65 | 98.83 | **98.91** | **99.01** |
| | NN freq. | 53.29 | 58.82 | 60.41 | 62.31 | 63.50 |
| `PDD-GCCA` ($\ell_{2,1}$) | AROC | 98.14 | **98.66** | **98.84** | **98.91** | **99.01** |
| | NN freq. | 53.45 | 58.92 | 60.50 | **62.36** | **63.52** |
| `LasCCA` | AROC | 96.31 | 96.71 | 96.53 | 96.44 | 96.64 |
| | NN freq. | 44.89 | 45.99 | 43.97 | 41.21 | 42.69 |
| `DisCCA` | AROC | 97.73 | 97.56 | 97.50 | 91.88 | 91.81 |
| | NN freq. | 47.78 | 42.99 | 41.38 | 17.87 | 15.31 |
| `MVLSA` | AROC | 93.60 | 93.42 | 92.11 | 91.35 | 91.83 |
| | NN freq. | 20.38 | 18.99 | 15.94 | 15.03 | 15.35 |

## VI. CONCLUSION

In this work, the regularized SUMCOR GCCA problem was considered. A highly scalable algorithmic framework that is based on the penalty-dual decomposition (PDD) strategy was proposed to address this challenging optimization problem. The proposed `PDD-GCCA` algorithm can easily incorporate many different regularizers to enforce structural canonical components, and thus is very flexible. It also admits lightweight updates and low memory complexity when handling large data. The proposed algorithmic framework is friendly to parallel computing – the variable splitting and dual decomposition nature of the algorithmic structure can easily facilitate distributed implementation with limited communication overhead. Despite the hardness of analyzing convergence properties of primal-dual optimization involving nonconvex constraints, we have been able to prove that the algorithm features KKT convergence assurance. Simulations on synthetic large-scale data have been employed to demonstrate the effectiveness of the proposed algorithm. Real data experiments on very large-scale cross language retrieval tasks, using data from the European Parliament, were used to showcase the advantages of the proposed algorithm in real world large scale applications.

## APPENDIX A
## PROOF OF PROPOSITION 1

Robinson's condition is a structural property of an optimization problem in the sense that it characterizes the geometry of the feasible set at a single point $x_0$. A system that satisfies the Robinson's condition at point $x_0$ is called metrically regular at that point (the notions of metric regularity and Robinson's condition have been proven equivalent) [50]. Specifically, consider the following system:

$$\begin{aligned} \theta(\boldsymbol{x}) &\in \mathcal{Y}_0 \\ \boldsymbol{x} &\in \mathcal{X}_0 \end{aligned} \quad (19)$$

where $\theta : \mathbb{R}^n \to \mathbb{R}^m$ is a continuously differentiable function, $\mathcal{Y}_0$ is a closed convex set in $\mathbb{R}^m$, and $\mathcal{X}_0$ is a closed convex set in $\mathbb{R}^n$.

**Definition 1** *[50] A system is called metrically regular at a point $\boldsymbol{x}_0 \in \mathcal{X}$ if there exist $\epsilon > 0$ and $C$ such that for all $\tilde{\boldsymbol{x}}$ and $\tilde{\boldsymbol{u}}$ satisfying $\|\tilde{\boldsymbol{x}} - \boldsymbol{x}_0\| \leq \epsilon$ and $\|\tilde{\boldsymbol{u}}\| \leq \epsilon$ we can find $\boldsymbol{x}_R \in \mathcal{X}_0$ satisfying the inclusion:*

$$\theta(\boldsymbol{x}_R) - \tilde{\boldsymbol{u}} \in \mathcal{Y}_0 \quad (20)$$

*and such that*

$$\|\tilde{\boldsymbol{x}} - \boldsymbol{x}_0\| \leq C(\text{dist}(\tilde{\boldsymbol{x}}, \mathcal{X}_0) + \text{dist}(\theta(\tilde{\boldsymbol{x}}) - \tilde{\boldsymbol{u}}, \mathcal{Y}_0)), \quad (21)$$

*where $\text{dist}(\theta(\tilde{\boldsymbol{x}}), \mathcal{X}_0)$ measures the Euclidean distance between point $\tilde{\boldsymbol{x}}$ and set $\mathcal{X}_0$.*

In simple words a system is called metrically regular at a point $\boldsymbol{x}_0$ if the distance of a perturbed point $\tilde{\boldsymbol{x}}$ from the solution $\boldsymbol{x}_R$ of a perturbed system is proportional to the violation of the two constraints. Although it is in general cumbersome to check metric regularity of a system, in case of equality constraints it becomes easier due to the following property.

**Property 1** *[50]: In the special case of equality constraints $\theta(\boldsymbol{x}) = 0$, when $\mathcal{X}_0 = \mathbb{R}^n$ and $\mathcal{Y}_0 = \{0\}$, metric regularity is equivalent to the linear independence of the gradients of the constraints $(\nabla \theta_i(\boldsymbol{x}_0), \; i = 1, \dots, m)$*

Hence, for continuously differentiable equality constraints, Robinson's condition is equivalent to the well known *Linear Independence Constraint Qualification* (LICQ) [51]. One way to prove the above property is by showing that the rank of the Jacobian $\nabla \theta(\boldsymbol{x}_0)$, defined by the set of equality constraints $\theta(\boldsymbol{x}) = 0$, is equal to m (the dimension of the co-domain of $\theta(\boldsymbol{x})$).

In this line of work, the interest lies upon checking Robinson's condition at a limit point $\{\boldsymbol{Q}_i^\star, \boldsymbol{G}_i^\star\}_{i=1}^I$ of the sequence $\{\boldsymbol{Q}_i^{(r)}, \boldsymbol{G}_i^{(r)}\}_{i=1}^I$ generated by Algorithm 1, for Problem (11).

The feasible set of the SUMCOR GCCA problem, written as in Eqs. (11b)-(11c), is presented below:

$$\boldsymbol{G}_i^T\boldsymbol{G}_i = \boldsymbol{I}_K, \quad i=1,\ldots,I \quad (22a)$$
$$\boldsymbol{X}_i\boldsymbol{Q}_i = \boldsymbol{G}_i, \quad i=1,\ldots,I \quad (22b)$$
$$\boldsymbol{X}_i \in \mathbb{R}^{L \times M_i}, \boldsymbol{Q}_i \in \mathbb{R}^{M_i \times K}, \boldsymbol{G}_i \in \mathbb{R}^{L \times K} \quad (22c)$$

Since the above system only involves equality constraints, proving that Robinson's condition holds for a limit point of PDD-GCCA algorithm boils down to proving that the gradients of each constraint, defined by equation, evaluated at a limit point $x_0$ are linearly independent.

The process of proving linear independence between the constraint gradients is sketched as follows. First, a unified optimization variable is defined that concatenates the variables $\boldsymbol{Q}_i$, $\boldsymbol{G}_i$, for $i=1,\ldots I$. Second, the matrix equality constraints are disassembled in a vector-scalar form and the gradient of each constraint is computed with respect to the single optimization variable. The gradients are aggregated in a matrix, which is equal to the Jacobian transpose. The final step verifies that the gradients of the constraints are independent, i.e. the Jacobian is full row rank, under very mild assumptions (if these assumptions do not hold, (G)CCA does not make sense).

Now, let $\boldsymbol{g}_i = \text{vec}(\boldsymbol{G}_i)$ and $\boldsymbol{q}_i = \text{vec}(\boldsymbol{Q}_i)$ for $i=1,\ldots,I$, where $\text{vec}(\cdot)$ denotes the column-wise vectorization operator of a matrix. Then the unified optimization variable $\boldsymbol{x} \in \mathbb{R}^{K(IL+\sum_{i=1}^I M_i)}$ is defined as:

$$\boldsymbol{x} = [\boldsymbol{g}_1;\ldots;\boldsymbol{g}_I;\boldsymbol{q}_1;\ldots;\boldsymbol{q}_I] \quad (23)$$

Considering the first set of equality constraints of problem (11), we have:

$$\boldsymbol{G}_i^T\boldsymbol{G}_i = \boldsymbol{I}_K, \quad i=1,\ldots,I \quad (24)$$

It is easy to observe that for each $i$, equation (24) corresponds to a $\frac{K(K+1)}{2}$ set of quadratic and linear scalar equations i.e:

$$\omega_i^1(\boldsymbol{x}) = \boldsymbol{G}_i(:,1)^T\boldsymbol{G}_i(:,1) - 1 = 0$$
$$\omega_i^2(\boldsymbol{x}) = \boldsymbol{G}_i(:,1)^T\boldsymbol{G}_i(:,2) = 0$$
$$\vdots$$
$$\omega_i^K(\boldsymbol{x}) = \boldsymbol{G}_i(:,1)^T\boldsymbol{G}_i(:,K) = 0$$
$$\omega_i^{K+1}(\boldsymbol{x}) = \boldsymbol{G}_i(:,2)^T\boldsymbol{G}_i(:,2) - 1 = 0$$
$$\omega_i^{K+2}(\boldsymbol{x}) = \boldsymbol{G}_i(:,2)^T\boldsymbol{G}_i(:,3) = 0$$
$$\vdots$$
$$\omega_i^{2K-1}(\boldsymbol{x}) = \boldsymbol{G}_i(:,2)^T\boldsymbol{G}_i(:,K) = 0$$
$$\vdots$$
$$\omega_i^{\frac{K(K+1)}{2}}(\boldsymbol{x}) = \boldsymbol{G}_i(:,K)^T\boldsymbol{G}_i(:,K) = 0$$

For each $i$ define a $LK \times \frac{K(K+1)}{2}$ matrix $\boldsymbol{\Omega}_i$ as:

$$\boldsymbol{\Omega}_i(:,j) = \nabla_{\boldsymbol{g}_i}\omega_i^j$$

$\boldsymbol{\Omega}_i$ is the Jacobian transpose of $\omega_i$ with respect to $\boldsymbol{g}_i$. So each column $j$ of matrix $\boldsymbol{\Omega}_i$ represents the gradient of $\omega_i^j$ with respect to $\boldsymbol{g}_i$. Matrix $\boldsymbol{\Omega}_i$ takes the form of Eq. (25) on the top of the next page, and is full column rank as long as $L \geq \frac{K+1}{2}$, which is the case in practice and $\boldsymbol{G}_i$ does not contain any zero columns. To be more precise, observe that the first $K$ columns of matrix $\boldsymbol{\Omega}_i$. These columns form a full column rank submatrix, if $\boldsymbol{G}_i$ does not contain any zero columns, since any column contains a non-zero vector at a unique place where the rest of the columns are zero. Now, observe that the next $K-1$ columns. They also form a full column rank submatrix for the same reason as before. Moreover the first set of $K$ columns and the second set of $K-1$ columns are linearly independent, since the upper $L \times K$ part of the first submatrix contains non-zero columns, whereas the upper $L \times (K-1)$ part of the second submatrix is an all-zero matrix. Following the same argument, we conclude that $\boldsymbol{\Omega}_i$ has full column rank, if it is square or tall, i.e., $L \geq \frac{K+1}{2}$, and $\boldsymbol{G}_i$ does not contain any zero columns. Note that since the proposed PDD-GCCA keeps the set of equalities (24) as constraints (does not relax this set of constraints), the produced $\boldsymbol{G}_i$'s are orthogonal and never contain zero columns.

Now, consider the second set of equality constraints of problem (11):

$$\boldsymbol{X}_i\boldsymbol{Q}_i = \boldsymbol{G}_i, \quad i=1,\ldots,I \quad (26)$$

and let $\boldsymbol{e}_j$ the j-th unit vector ($\boldsymbol{e}_j$ has only one nonzero element at the j-th place and is equal to 1). We observe that for each $i$ (26) corresponds to a $LK$ set of linear scalar equations i.e:

$$\phi_i^1(\boldsymbol{x}) = \boldsymbol{X}_i(1,:)\boldsymbol{Q}_i(:,1) - \boldsymbol{e}_1^T\boldsymbol{G}_i(:,1) = 0$$
$$\phi_i^2(\boldsymbol{x}) = \boldsymbol{X}_i(2,:)\boldsymbol{Q}_i(:,1) - \boldsymbol{e}_2^T\boldsymbol{G}_i(:,1) = 0$$
$$\vdots$$
$$\phi_i^L(\boldsymbol{x}) = \boldsymbol{X}_i(L,:)\boldsymbol{Q}_i(:,1) - \boldsymbol{e}_L^T\boldsymbol{G}_i(:,1) = 0$$
$$\phi_i^{L+1}(\boldsymbol{x}) = \boldsymbol{X}_i(1,:)\boldsymbol{Q}_i(:,2) - \boldsymbol{e}_1^T\boldsymbol{G}_i(:,2) = 0$$
$$\phi_i^{L+2}(\boldsymbol{x}) = \boldsymbol{X}_i(2,:)\boldsymbol{Q}_i(:,2) - \boldsymbol{e}_2^T\boldsymbol{G}_i(:,2) = 0$$
$$\vdots$$
$$\phi_i^{2L}(\boldsymbol{x}) = \boldsymbol{X}_i(L,:)\boldsymbol{Q}_i(:,2) - \boldsymbol{e}_L^T\boldsymbol{G}_i(:,2) = 0$$
$$\vdots$$
$$\phi_i^{(K-1)L+1}(\boldsymbol{x}) = \boldsymbol{X}_i(1,:)\boldsymbol{Q}_i(:,K) - \boldsymbol{e}_1^T\boldsymbol{G}_i(:,K) = 0$$
$$\phi_i^{(K-1)L+2}(\boldsymbol{x}) = \boldsymbol{X}_i(2,:)\boldsymbol{Q}_i(:,K) - \boldsymbol{e}_2^T\boldsymbol{G}_i(:,K) = 0$$
$$\vdots$$
$$\phi_i^{KL}(\boldsymbol{x}) = \boldsymbol{X}_i(L,:)\boldsymbol{Q}_i(:,K) - \boldsymbol{e}_L^T\boldsymbol{G}_i(:,K) = 0$$

For each $i$ define a $M_iK \times LK$ matrix $\boldsymbol{\Phi}_i$ as:

$$\boldsymbol{\Phi}_i(:,j) = \nabla_{\boldsymbol{q}_i}\phi_i^j \quad (27)$$

which is the Jacobian transpose of $\phi_i$ with respect to $\boldsymbol{q}_i$, so each column $j$ of matrix $\boldsymbol{\Phi}_i$ represents the gradient of $\phi_i^j$ with respect to $\boldsymbol{q}_i$. Matrix $\boldsymbol{\Phi}_i$ takes the following form:

$$\boldsymbol{\Phi}_i = \begin{bmatrix} \boldsymbol{X}_i^T & \boldsymbol{0} & \ldots & \boldsymbol{0} \\ \boldsymbol{0} & \boldsymbol{X}_i^T & \ldots & \boldsymbol{0} \\ \vdots & \vdots & \ddots & \vdots \\ \boldsymbol{0} & \boldsymbol{0} & \ldots & \boldsymbol{X}_i^T \end{bmatrix} \quad (28)$$

$$\boldsymbol{\Omega}_i = \begin{bmatrix} \overbrace{2\boldsymbol{G}_i(:,1) \quad \boldsymbol{G}_i(:,2) \quad \boldsymbol{G}_i(:,3) \quad \ldots \quad \boldsymbol{G}_i(:,K)}^{K} & \overbrace{\boldsymbol{0} \quad \boldsymbol{0} \quad \ldots \quad \boldsymbol{0}}^{K-1} & \ldots & \overbrace{\boldsymbol{0}}^{1} \\ 0 & \boldsymbol{G}_i(:,1) & 0 & \ldots & 0 & 2\boldsymbol{G}_i(:,2) & \boldsymbol{G}_i(:,3) & \ldots & \boldsymbol{G}_i(:,K) & \ldots & 0 \\ 0 & 0 & \boldsymbol{G}_i(:,1) & \ldots & 0 & 0 & \boldsymbol{G}_i(:,2) & \ldots & 0 & \ldots & 0 \\ \vdots & \vdots & \vdots & \ddots & \vdots & \vdots & \vdots & \ddots & \vdots & \vdots & \vdots \\ 0 & 0 & 0 & \ldots & \boldsymbol{G}_i(:,1) & 0 & 0 & \ldots & \boldsymbol{G}_i(:,2) & \ldots & 2\boldsymbol{G}_i(:,K) \end{bmatrix} \quad (25)$$

Moreover the the gradient of $\phi_i^j$ with respect to $\boldsymbol{q}_i$ is the vector $-\boldsymbol{e}_j \in \mathbb{R}^{KL \times 1}$. Consequently the Jacobian of $\phi_i$ with respect to $\boldsymbol{q}_i$ is the negative of a $KL$ identity matrix, i.e. $-\boldsymbol{I}_{KL}$. Note that $\boldsymbol{\Phi}_i$ is not full rank in general. This is not an issue however as it is shown later. Now let

$$\begin{pmatrix} \theta_1(\boldsymbol{x}) \\ \vdots \\ \theta_{\frac{K(K+1)}{2}}(\boldsymbol{x}) \\ \theta_{\frac{K(K+1)}{2}+1}(\boldsymbol{x}) \\ \vdots \\ \theta_{K(K+1)}(\boldsymbol{x}) \\ \vdots \\ \theta_{\frac{IK(K+1)}{2}}(\boldsymbol{x}) \end{pmatrix} = \begin{pmatrix} \omega_1^1(\boldsymbol{x}) \\ \vdots \\ \omega_1^{\frac{K(K+1)}{2}}(\boldsymbol{x}) \\ \omega_2^1(\boldsymbol{x}) \\ \vdots \\ \omega_2^{\frac{K(K+1)}{2}}(\boldsymbol{x}) \\ \vdots \\ \omega_I^{\frac{K(K+1)}{2}}(\boldsymbol{x}) \end{pmatrix}$$

and

$$\begin{pmatrix} \theta_{\frac{IK(K+1)}{2}+1}(\boldsymbol{x}) \\ \vdots \\ \theta_{\frac{IK(K+1)}{2}+KL}(\boldsymbol{x}) \\ \theta_{\frac{IK(K+1)}{2}+KL+1}(\boldsymbol{x}) \\ \vdots \\ \theta_{\frac{IK(K+1)}{2}+2KL}(\boldsymbol{x}) \\ \vdots \\ \theta_{\frac{IK(K+1)}{2}+IKL}(\boldsymbol{x}) \end{pmatrix} = \begin{pmatrix} \phi_1^1(\boldsymbol{x}) \\ \vdots \\ \phi_1^{KL}(\boldsymbol{x}) \\ \phi_2^1(\boldsymbol{x}) \\ \vdots \\ \phi_2^{KL}(\boldsymbol{x}) \\ \vdots \\ \phi_I^{KL}(\boldsymbol{x}) \end{pmatrix}$$

then

$$\theta_i(\boldsymbol{x}) = 0, \quad i = 1, \ldots, \frac{IK(K+1)}{2} + IKL \quad (29)$$

is equivalent to the system of equations (22).
The set of gradients

$$\nabla_{\boldsymbol{x}} \theta_i(\boldsymbol{x}), \quad i = 1, \ldots, \frac{IK(K+1)}{2} + IKL \quad (30)$$

are represented by the columns of matrix $\boldsymbol{\Theta}$, which is the Jacobian transpose of $\theta(\boldsymbol{x})$:

$$\boldsymbol{\Theta} = \begin{bmatrix} \overbrace{\boldsymbol{\Omega}_1 \quad \boldsymbol{0} \quad \ldots \quad \boldsymbol{0}}^{IK(K+1)/2} & \overbrace{-\boldsymbol{I}_{KL} \quad \boldsymbol{0} \quad \ldots \quad \boldsymbol{0}}^{IKL} \\ \boldsymbol{0} & \boldsymbol{\Omega}_2 & \ldots & \boldsymbol{0} & \boldsymbol{0} & -\boldsymbol{I}_{KL} & \ldots & \boldsymbol{0} \\ \vdots & \vdots & \ddots & \vdots & \vdots & \vdots & \ddots & \vdots \\ \boldsymbol{0} & \boldsymbol{0} & \ldots & \boldsymbol{\Omega}_I & \boldsymbol{0} & \boldsymbol{0} & \ldots & -\boldsymbol{I}_{KL} \\ \boldsymbol{0} & \boldsymbol{0} & \ldots & \boldsymbol{0} & \boldsymbol{\Phi}_1 & \boldsymbol{0} & \ldots & \boldsymbol{0} \\ \boldsymbol{0} & \boldsymbol{0} & \ldots & \boldsymbol{0} & \boldsymbol{0} & \boldsymbol{\Phi}_2 & \ldots & \boldsymbol{0} \\ \vdots & \vdots & \ddots & \vdots & \vdots & \vdots & \ddots & \vdots \\ \boldsymbol{0} & \boldsymbol{0} & \ldots & \boldsymbol{0} & \boldsymbol{0} & \boldsymbol{0} & \ldots & \boldsymbol{\Phi}_I \end{bmatrix}$$

$\boldsymbol{\Theta}$ is full column rank under the following conditions:

$$\frac{1}{I} \sum_{i=1}^{I} M_i \geq \frac{K+1}{2} \quad (31a)$$

$$L \geq \frac{K+1}{2}, \quad (31b)$$

$$\boldsymbol{G}_i \text{ do not contain zero columns } \forall i. \quad (31c)$$

Since in any (G)CCA problem $M_i > K$, $\forall i$, (31a) is trivial and always satisfied. Furthermore for any (G)CCA problem the number of observations $L$ is always greater than the number of canonical components $K$, which makes (31b) trivial as well. Finally, the proposed `PDD-GCCA` algorithm always produces $\boldsymbol{G}_i$'s that do not contain zero columns, as explained earlier. Thus $\boldsymbol{\Theta}$ is full column rank and the set of gradients

$$\nabla_{\boldsymbol{x}} \theta_i(\boldsymbol{x}), \quad i = 1, \ldots, \frac{IK(K+1)}{2} + IKL \quad (32)$$

are linearly independent.

Consequently, under the conditions (31a), (31b) any set of points $\{\boldsymbol{Q}_i, \boldsymbol{G}_i\}_{i=1}^{I}$ generated by Algorithm 1 for Problem (11) is metrically regular and satisfies the Robinson's conditions.

REFERENCES

[1] H. Hotelling, "Relations between two sets of variates," *Biometrika*, vol. 28, no. 3/4, pp. 321–377, 1936.
[2] J. R. Kettenring, "Canonical analysis of several sets of variables," *Biometrika*, vol. 58, no. 3, pp. 433–451, 1971.
[3] J. D. Carroll, "Generalization of canonical correlation analysis to three or more sets of variables," in *Proceedings of the 76th annual convention of the American Psychological Association*, vol. 3, 1968, pp. 227–228.
[4] S. Bickel and T. Scheffer, "Multi-view clustering." in *ICDM*, vol. 4, 2004, pp. 19–26.
[5] S. M. Kakade and D. P. Foster, "Multi-view regression via canonical correlation analysis," in *Learning Theory*. Springer, 2007, pp. 82–96.

[6] M. A. Alam and Y.-P. Wang, "Identifying outliers using influence function of multiple kernel canonical correlation analysis," *arXiv preprint arXiv:1606.00113*, 2016.

[7] X. Fu, K. Huang, M. Hong, N. D. Sidiropoulos, and A. M.-C. So, "Scalable and flexible max-var generalized canonical correlation analysis via alternating optimization," *IEEE Trans. Signal Process.*, to appear, 2017.

[8] X. Fu, K. Huang, E. Papalexakis, H. Song, P. Talukdar, N. D. Sidiropoulos, C. Faloutsos, and T. Mitchell, "Efficient and distributed algorithms for large-scale generalized correlation analysis," in *Proc. ICDM 2016*. IEEE, 2016.

[9] P. Rastogi, B. Van Durme, and R. Arora, "Multiview lsa: Representation learning via generalized cca." in *HLT-NAACL*, 2015, pp. 556–566.

[10] R. Arora and K. Livescu, "Multi-view learning with supervision for transformed bottleneck features," in *Proc. ICASSP 2014*, 2014, pp. 2499–2503.

[11] J. C. Vasquez-Correa, J. R. Orozco-Arroyave, R. Arora, E. Nöth, N. Dehak, H. Christensen, F. Rudzicz, T. Bocklet, M. Cernak, H. Chinaei et al., "Multi-view representation learning via gcca for multimodal analysis of parkinson" s disease," in *Proceedings of 2017 IEEE International Conference on Acoustics, Speech, and Signal Processing (ICASSP 2017)*, no. EPFL-CONF-224545, 2017.

[12] E. Parkhomenko, D. Tritchler, J. Beyene et al., "Sparse canonical correlation analysis with application to genomic data integration," *Statistical Applications in Genetics and Molecular Biology*, vol. 8, no. 1, pp. 1–34, 2009.

[13] D. M. Witten and R. J. Tibshirani, "Extensions of sparse canonical correlation analysis with applications to genomic data," *Statistical applications in genetics and molecular biology*, vol. 8, no. 1, pp. 1–27, 2009.

[14] D. M. Witten, R. Tibshirani, and T. Hastie, "A penalized matrix decomposition, with applications to sparse principal components and canonical correlation analysis," *Biostatistics*, p. kxp008, 2009.

[15] R. Ge, C. Jin, S. M. Kakade, P. Netrapalli, and A. Sidford, "Efficient algorithms for large-scale generalized eigenvector computation and canonical correlation analysis," *arXiv preprint arXiv:1604.03930*, 2016.

[16] Z. Allen-Zhu and Y. Li, "Doubly accelerated methods for faster cca and generalized eigendecomposition," *arXiv preprint arXiv:1607.06017*, 2016.

[17] Y. Lu and D. P. Foster, "Large scale canonical correlation analysis with iterative least squares," in *NIPS*, 2014, pp. 91–99.

[18] W. Wang, J. Wang, and N. Srebro, "Globally convergent stochastic optimization for canonical correlation analysis," *arXiv preprint arXiv:1604.01870*, 2016.

[19] J. Rupnik, P. Skraba, J. Shawe-Taylor, and S. Guettes, "A comparison of relaxations of multiset cannonical correlation analysis and applications," *arXiv preprint arXiv:1302.0974*, 2013.

[20] Z. Ma, Y. Lu, and D. Foster, "Finding linear structure in large datasets with scalable canonical correlation analysis," in *ICML 2015*, 2015.

[21] S. Waaijenborg, P. C. Verselewel de Witt Hamer, and A. H. Zwinderman, "Quantifying the association between gene expressions and dna-markers by penalized canonical correlation analysis," *Statistical applications in genetics and molecular biology*, vol. 7, no. 1, 2008.

[22] G. H. Golub and H. Zha, *The canonical correlations of matrix pairs and their numerical computation*. Springer, 1995.

[23] X. Fu, K. Huang, M. Hong, N. D. Sidiropoulos, and A. M.-C. So, "Scalable and optimal generalized canonical correlation analysis via alternating optimization," *arXiv preprint arXiv:1605.09459*, 2016.

[24] P. Rastogi, B. Van Durme, and R. Arora, "Multiview LSA: Representation learning via generalized cca," in *NAACL*, 2015.

[25] L. Sun, S. Ji, and J. Ye, "Canonical correlation analysis for multilabel classification: A least-squares formulation, extensions, and analysis," *IEEE Trans. Patt. Analysis Machine Intel.*, vol. 33, no. 1, pp. 194–200, 2011.

[26] X. Chen, H. Liu, and J. G. Carbonell, "Structured sparse canonical correlation analysis," in *International Conference on Artificial Intelligence and Statistics*, 2012, pp. 199–207.

[27] A. Tenenhaus, C. Philippe, V. Guillemot, K.-A. Le Cao, J. Grill, and V. Frouin, "Variable selection for generalized canonical correlation analysis," *Biostatistics*, p. kxu001, 2014.

[28] C. Gao, Z. Ma, and H. H. Zhou, "An efficient and optimal method for sparse canonical correlation analysis," *arXiv preprint arXiv:1409.8565*, 2014.

[29] Q. Shi and M. Hong, "Penalty dual decomposition method with application in signal processing," in *Acoustics, Speech and Signal Processing (ICASSP), 2017 IEEE International Conference on*. IEEE, 2017, pp. 4059–4063.

[30] C. I. Kanatsoulis, X. Fu, N. D. Sidiropoulos, and M. Hong, "Large-scale regularized sumcor gcca via penalty-dual decomposition," submitted to *ICASSP 2018*.

[31] D. R. Hardoon, S. Szedmak, and J. Shawe-Taylor, "Canonical correlation analysis: An overview with application to learning methods," *Neural computation*, vol. 16, no. 12, pp. 2639–2664, 2004.

[32] A. Tenenhaus and M. Tenenhaus, "Regularized generalized canonical correlation analysis," *Psychometrika*, vol. 76, no. 2, pp. 257–284, 2011.

[33] L.-H. Zhang, L.-Z. Liao, and L.-M. Sun, "Towards the global solution of the maximal correlation problem," *Journal of Global Optimization*, vol. 49, no. 1, pp. 91–107, 2011.

[34] M. T. Chu and J. L. Watterson, "On a multivariate eigenvalue problem, part i: Algebraic theory and a power method," *SIAM Journal on scientific computing*, vol. 14, no. 5, pp. 1089–1106, 1993.

[35] J. R. Shewchuk, "An introduction to the conjugate gradient method without the agonizing pain," 1994.

[36] A. Beck and M. Teboulle, "A fast iterative shrinkage-thresholding algorithm for linear inverse problems," *SIAM journal on imaging sciences*, vol. 2, no. 1, pp. 183–202, 2009.

[37] N. Parikh and S. Boyd, "Proximal algorithms," *Foundations and Trends in optimization*, vol. 1, no. 3, pp. 123–231, 2013.

[38] Q. Shi and M. Hong, "Penalty dual decomposition method with application in signal processing," in *Acoustics, Speech and Signal Processing (ICASSP), 2017 IEEE International Conference on*. IEEE, 2017, pp. 4059–4063.

[39] D. P. Bertsekas, *Nonlinear programming*. Athena Scientific, 1999.

[40] M. Razaviyayn, M. Hong, and Z.-Q. Luo, "A unified convergence analysis of block successive minimization methods for nonsmooth optimization," *SIAM Journal on Optimization*, vol. 23, no. 2, pp. 1126–1153, 2013.

[41] J. Tranter, N. D. Sidiropoulos, X. Fu, and A. Swami, "Fast unit-modulus least squares with applications in beamforming," *IEEE Transactions on Signal Processing*, vol. 65, no. 11, pp. 2875–2887, 2017.

[42] L. Ballesteros and W. B. Croft, "Phrasal translation and query expansion techniques for cross-language information retrieval," in *ACM SIGIR Forum*, vol. 31, no. SI. ACM, 1997, pp. 84–91.

[43] J.-Y. Nie, M. Simard, P. Isabelle, and R. Durand, "Cross-language information retrieval based on parallel texts and automatic mining of parallel texts from the web," in *Proceedings of the 22nd annual international ACM SIGIR conference on Research and development in information retrieval*. ACM, 1999, pp. 74–81.

[44] W. Y. Zou, R. Socher, D. Cer, and C. D. Manning, "Bilingual word embeddings for phrase-based machine translation," in *Proceedings of the 2013 Conference on Empirical Methods in Natural Language Processing*, 2013, pp. 1393–1398.

[45] P. Koehn, "Europarl: A parallel corpus for statistical machine translation," in *MT summit*, vol. 5, 2005, pp. 79–86.

[46] K. Weinberger, A. Dasgupta, J. Langford, A. Smola, and J. Attenberg, "Feature hashing for large scale multitask learning," in *Proceedings of the 26th Annual International Conference on Machine Learning*. ACM, 2009, pp. 1113–1120.

[47] P. Mineiro and N. Karampatziakis, "A randomized algorithm for cca," *arXiv preprint arXiv:1411.3409*, 2014.

[48] A. P. Bradley, "The use of the area under the roc curve in the evaluation of machine learning algorithms," *Pattern recognition*, vol. 30, no. 7, pp. 1145–1159, 1997.

[49] D. Chu, L.-Z. Liao, M. K. Ng, and X. Zhang, "Sparse canonical correlation analysis: new formulation and algorithm," *IEEE transactions on pattern analysis and machine intelligence*, vol. 35, no. 12, pp. 3050–3065, 2013.

[50] A. P. Ruszczyński, *Nonlinear optimization*. Princeton university press, 2006, vol. 13.

[51] D. W. Peterson, "A review of constraint qualifications in finite-dimensional spaces," *SIAM Review*, vol. 15, no. 3, pp. 639–654, 1973.